\documentclass{ecai}
\usepackage{times}
\usepackage{graphicx}
\usepackage{latexsym}

\usepackage{url}
\usepackage{amsmath}
\usepackage{amssymb}
\usepackage{multirow}
\usepackage{color}
\usepackage{subfigure}
\usepackage{verbatim}
\usepackage{enumitem}
\usepackage{pifont}
\usepackage{mathrsfs}
\usepackage{booktabs}
\usepackage[numbers,sort&compress]{natbib}
\usepackage{acronym}
\usepackage{lipsum}

\usepackage{threeparttable}
 
\acrodef{SPDS}{Structured Persona-oriented Dialogue Systems}
\acrodef{UPDS}{Unstructured Persona-oriented Dialogue Systems}
\acrodef{PEE}{Persona Exploration and Exploitation}
\acrodef{P-Match}{Persona-oriented Matching}
\acrodef{P-BoWs}{Persona-oriented Bag-of-Words}

\makeatletter
\def\blfootnote{\gdef\@thefnmark{}\@footnotetext}
\makeatother

\begin{document}

\title{A Neural Topical Expansion Framework for \\Unstructured Persona-oriented Dialogue Generation}

\author{Minghong Xu\institute{Shandong University, China, email: minghongxu.sd@outlook.com} 
\and Piji Li\institute{Tencent AI Lab, China, email: pijili@tencent.com} \ $^\ast$
\and Haoran Yang\institute{The Chinese University of Hong Kong, email: hryang@se.cuhk.edu.hk}  
\and Pengjie Ren\institute{University of Amsterdam, The Netherlands, email: p.ren@uva.nl} \\ 
\and Zhaochun Ren\institute{Shandong University, China, email: zhaochun.ren@sdu.edu.cn} \ $^\ast$
\and Zhumin Chen\institute{Shandong University, China, email: chenzhumin@sdu.edu.cn} 
\and Jun Ma\institute{Shandong University, China, email: majun@sdu.edu.cn}}

\maketitle

\begin{abstract}
\ac{UPDS} has been demonstrated effective in generating persona consistent responses by utilizing predefined natural language user persona descriptions (e.g., ``I am a vegan'').
However, the predefined user persona descriptions are usually short and limited to only a few descriptive words, which makes it hard to correlate them with the dialogues.
As a result, existing methods either fail to use the persona description or use them improperly when generating persona consistent responses.
To address this, we propose a neural topical expansion framework, namely \ac{PEE}, which is able to extend the predefined user persona description with semantically correlated content before utilizing them to generate dialogue responses.
\ac{PEE} consists of two main modules: persona exploration and persona exploitation.
The former learns to extend the predefined user persona description by mining and correlating with existing dialogue corpus using a variational auto-encoder (VAE) based topic model.
The latter learns to generate persona consistent responses by utilizing the predefined and extended user persona description.
In order to make persona exploitation learn to utilize user persona description more properly, we also introduce two persona-oriented loss functions: \ac{P-Match} loss and \ac{P-BoWs} loss which respectively supervise persona selection in encoder and decoder.
Experimental results show that our approach outperforms state-of-the-art baselines, in terms of both automatic and human evaluations. 
\end{abstract}

\blfootnote{$^\ast$ Piji Li and Zhaochun Ren are corresponding authors.}


\section{Introduction}
Persona-oriented dialogue systems have attracted an increasing attention as they can generate persona consistent  responses~\cite{li,DBLP:journals/corr/abs-1810-08717,DBLP:journals/corr/abs-1809-01984,DBLP:journals/corr/abs-1811-04604}. 
Existing persona-oriented dialogue systems can be classified into two categories: \acf{SPDS}~\cite{P17,ijcai2018-595,aaai2020} and \acf{UPDS}~\cite{P18-1205,Exploiting}. 
The former directly uses structured user persona descriptions in the form of key-value pairs (e.g., $\left \langle SEX, M \right \rangle$,$\left \langle AGE, 18 \right \rangle$), whereas the latter mines user persona descriptions from natural language utterances (e.g., ``\textit{I like music.}'', ``\textit{I like the guitar.}'', ``\textit{I am a vegan.}''). 
In this work, we focus on \ac{UPDS}. 

\begin{table}
    \centering
    \caption{An example of unstructured persona-oriented dialogue system.}
    \resizebox{0.9\columnwidth}{!}{\begin{tabular}{c|p{6.5cm}}
        \toprule
                &1. I like music. \\
        Personas for &2. I like to skateboard. \\ 
        Speaker B    &3. I like the guitar. \\
                &4. I am a \textbf{vegan}. \\
        \midrule 
                &\textbf{A}(u$_1$): Wanna come over and watch the godfather? \\
                &\textbf{B}(u$_2$): I do not have a car, I have a skateboard. \\ 
                &\textbf{A}(u$_3$): You can skateboard over. I do not live too far. I have candy and soda to share. \\
       Dialogue &\textbf{B}(u$_4$): No thanks, I do not eat any animal products. \\ 
                &\textbf{A}(u$_5$): I promise there are no \textbf{animal products} in my \textbf{candy} and soda. \\
                &\textbf{B}(u$_6$): Most \textbf{candy} has some form of \textbf{dairy}. As a \textbf{vegan} I can not have that. \\
        \bottomrule
    \end{tabular}}
    \label{table:example_for_persona_expansion}
    \vspace{-3mm}
\end{table}
Recently, there have been some studies which utilize the predefined user persona descriptions to generate persona-oriented responses \cite{P18-1205, Exploiting,lin2019personalizing}.
However, the given user persona descriptions are mostly short and limited to only a few descriptive words. 
As a result, the existing methods have a hard time utilizing the user persona descriptions when generating responses. 
On the one hand, they might fail to use user persona descriptions. 
For example, the generative profile memory network proposed in \cite{P18-1205} simply attends over encoded persona description in decoder. It generates response ``I have a lot of candies. I am not sure.'' for the case in Table~\ref{table:example_for_persona_expansion} without considering user persona. 
On the other hand, they cannot use user persona descriptions properly sometimes.
For example, the persona-CVAE proposed in \cite{Exploiting} uses force decoding strategy to copy persona description. 
It generates response  ``I like to skateboard. What are your hobbies?'' for the case in Table~\ref{table:example_for_persona_expansion} which use the selected persona improperly and seriously affects its quality.
The reason is that with the limited descriptive words, it is hard for these models to understand and correlate the user persona descriptions when generating responses.
We argue that this could be alleviated by extending the predefined user persona descriptions with semantically correlated content.
As shown in Table~\ref{table:example_for_persona_expansion},
the target is to generate the last utterance (u$_6$) based on the given persona descriptions and historical utterances (u$_1$-u$_5$).
One of the user persona descriptions for Speaker B is ``I am a vegan''. 
However, only using this user persona description is not enough to generate huaman-like response (u$_6$) because ``vegan'' and ``candy'' are not directly related. 
In order to generate u$_6$, we need to take the following content into consideration simultaneously: (1) the word ``vegan'' in B's user persona description; (2) the semantic correlation between ``vegan'' and ``dairy''; (3) speaker A mentioned ``animal products'' and ``candy'' in the query utterance; (4) the correlation among ``dairy'', ``animal products'', and ``candy''. 

In this work, we propose a neural topical expansion framework, namely \textbf{P}ersona \textbf{E}xploration and \textbf{E}xploitation (PEE), which is able to extend the predefined user persona descriptions with semantically correlated content before utilizing them to generate dialogue responses.
\ac{PEE} consists of two main modules: persona exploration and persona exploitation.
The former learns to extend the predefined user persona descriptions by mining and correlating with existing dialogue corpus.
Specifically, we employ a VAE-based topic model to conduct the unsupervised semantic modeling and extend persona-related words by semantic matching.
The latter learns to generate persona consistent responses by utilizing the predefined and extended persona information.
Specifically, we design a mutual-reinforcement multi-hop memory retrieval mechanism which retrieves information from two types (predefined and extended) of personas by considering their mutual influence.
Furthermore, in order to make persona exploitation learn to utilize user persona descriptions more properly, we also introduce two persona-oriented loss functions: \ac{P-Match} loss and \ac{P-BoWs} loss.
P-Match loss supervises the choice of predefined persona sentences in encoder. P-BoWs loss supervises to generate more persona-related words in decoder.

The main contributions of this paper are as follows: 
\begin{itemize}[nosep]
  \item We propose a persona exploration and exploitation (\textbf{PEE}) framework which can explore and exploit persona information to generate informative persona-oriented responses.
  \item We employ an VAE-based topic model to conduct the efficient unsupervised semantic learning for external persona information mining and distillation.
  \item We propose two learning strategies for persona exploitation: a mutual-reinforcement multi-hop memory retrieval mechanism and two persona-oriented loss functions. 
\end{itemize}


\section{Realted work}
As a challenging task in the area of natural language processing, open-domain dialogue system has attracted great attention of researchers recently~\cite{vinyals2015neural,serban2016hierarchical,ThinkingGlobally,refnet}.
But there are still some limitations and challenges in this area. Among the many issues, the lack of consistency is one of the most challenging difficulties.
Therefore, persona-based dialogue system has been proposed to generate persona consistent and human-like responses~\cite{Yang:2017:PRG:3077136.3080706,DBLP:journals/corr/JoshiMF17,Mo2018PersonalizingAD,urbanek2019light,DBLP:journals/corr/abs-1905-01992}.
~\citet{li} learn a user embedding to represent persona implicitly for each user without using explicitly persona information.
Then, researchers model user embedding with explicit persona information to generate responses. According to the format of persona information, those methods can be classified into two categories: structured Persona-oriented Dialogue Systems (SPDS) and Unstructured Persona-oriented Dialogue Systems (UPDS).

In SPDS, \citet{wang-etal-2017-group} group users according to the gender attribute and the dialogue features in the same group can be shared.
\citet{ijcai2018-595} endow the user with explicit structured persona information (a key-value table) and design a profile detection module to select a persona information and inject it to the decoding process. \citet{DBLP:journals/corr/abs-1811-04604} encode user persona description into distributed embeddings and take advantage of conversation history from other users with similar profiles, their model can adopt different recommendation policy based on the user profile. 
Due to the lack of large scale persona-labelled dataset, \citet{P17} introduce a dataset where persona information is formulated to key-value pairs from dialogue content and they devise two technique to  capture and address trait-related information. 
In UPDS, \citet{P18-1205} contribute a persona-chat dataset with natural sentences persona information and they propose a generative profile memory network to incorporate persona information into responses. 
\citet{lin2019personalizing} model learning different personas as different tasks via meta-learning algorithm without using explicit persona information since dialogue itself can reflect some persona information. Through this way, their model can generate personalised responses by leveraging only a few dialogue samples instead of human-designed persona descriptions. 
To generate diverse and sustainable conversations, \citet{Exploiting} propose a memory-augmented architecture to exploit persona information and utilized a conditional variational autoencoder which can address the one-to-many generation problem. 

Prior studies are trained purely on predefined persona corpus, but the limited information leads to generating uninformative responses.
Different from them, we employ VAE-based topic model to extend persona information and propose two strategies (a mutual-reinforcement multi-hop memory retrieval mechanism and two persona-oriented loss functions) to integrate persona information to responses.


\section{Method}

\subsection{Overview}

\begin{figure*}[!t]
 \centering
  \includegraphics[width=0.87\linewidth]{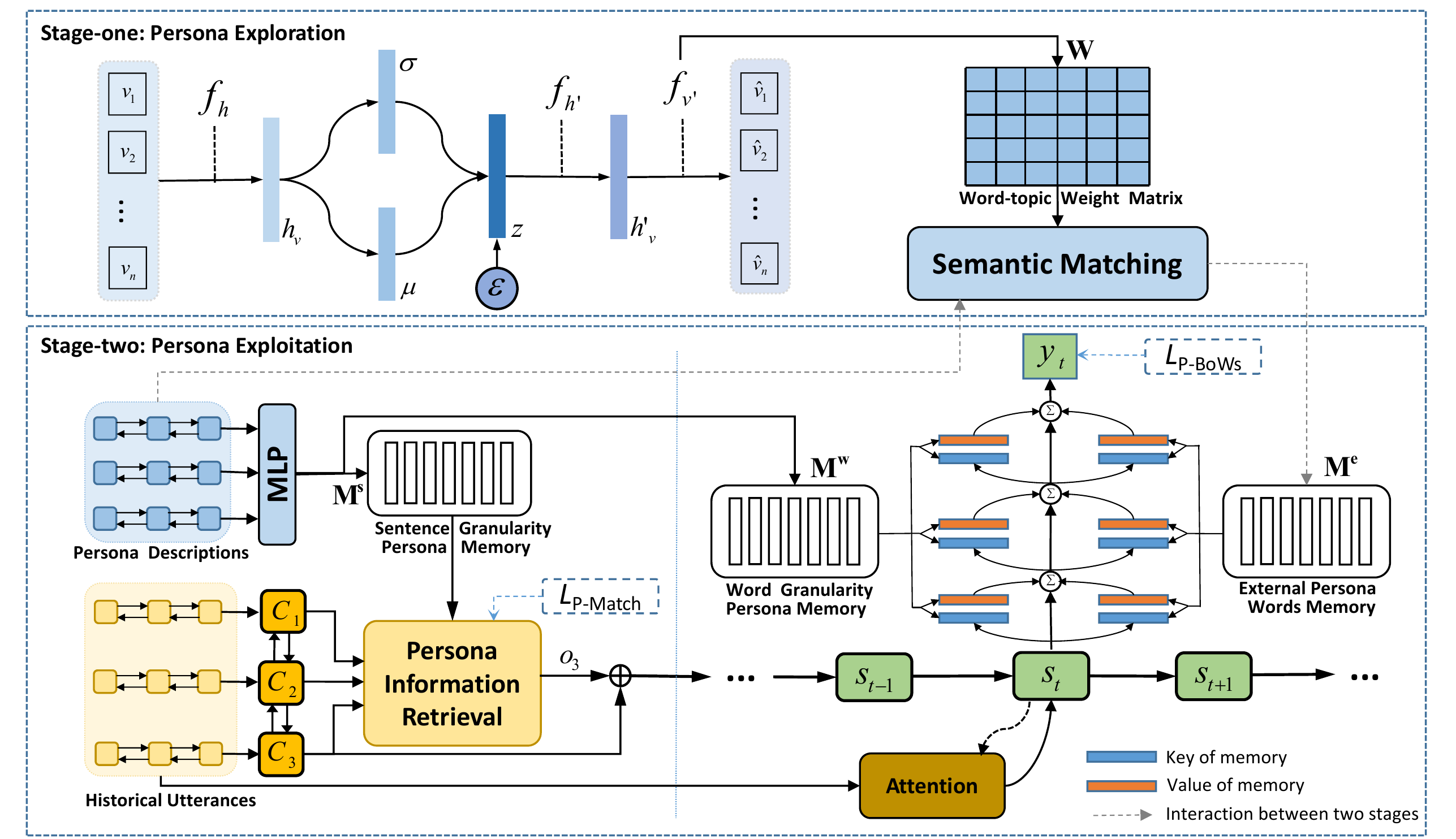}
  \caption{An overview of our PEE framework. It consists of two stages: persona exploration and persona exploitation (multi-source sequence encoder, persona information retrieval and persona-oriented response decoder).}
  \label{fig:overview_of_our_model}
  \vspace{-3mm}
\end{figure*}

We assume that a conversation is conducted between two users. 
Given a target user, we denote the user's persona descriptions as $P = (P_1, P_2, \ldots, P_{n_p})$. Each persona sentence $P_j$ is formulated as $P_j=(p^j_1, p^j_2, \ldots, p^j_{l_p})$, where $p_i^j$ refers to a word.
Suppose there are already $k$ turns in a dialogue, so we have historical utterances as $X = (X_1, X_2, \ldots, X_{k})$, where each utterance $X_{i}$ is depicted as $X_i=(x^i_1,x^i_2,\ldots,x^i_{l_x})$ and $x^i_j$ denotes a word.
Accordingly, \textit{unstructured persona-oriented dialogue generation} aims to predict the ($k$+$1$)-th utterance, i.e., the response $Y=(y_1,y_2,\ldots,y_{l_y})$, according to the predefined persona descriptions $P$ and the historical utterances $X$:
\begin{equation}
\setlength{\abovedisplayskip}{3pt}
\setlength{\belowdisplayskip}{3pt}
    \small
    p(Y|X, P) = \prod_{i=1}^{l_y}p\left( y_i | X,P,y_1, \ldots , y_{i-1} \right).
\end{equation}

As illustrated in Figure~\ref{fig:overview_of_our_model}, our PEE framework mainly consists of two stages: persona exploration and persona exploitation.
Persona exploration employs a VAE-based topic model to conduct the unsupervised semantic modeling and obtains topic-relevant word representations. Then it extends persona-related words by semantic matching based on the predefined persona descriptions.
Persona exploitation contains three components:
(1) multi-source sequence encoder, which encodes predefined persona descriptions into two kinds of key-value memories and encodes historical utterances into hidden vectors;
(2) persona information retrieval, which selects predefined persona descriptions based on historical utterances and considers the impact of personalized information involved in the history;
(3) persona-oriented response decoder, which exploits the predefined and explored external persona information to generate responses based on the specially designed mutual-reinforcement multi-hop memory retrieval mechanism.
Moreover, two new optimization objectives, persona-oriented matching loss (P-Match) and persona-oriented bag-of-words loss (P-BoWs), are proposed to impel our model to exploit the persona information more precisely.
We will introduce the technical details in the following sections.

\subsection{Persona Exploration}
\label{section2.3}

Based on the predefined persona descriptions, the target of the persona exploration stage is to extend more persona-related words. 
Therefore the key is to investigate an effective method for the semantic learning of words. 
The extended persona words must lie in the same topic with the predefined persona information, so as to guarantee the topic consistence of the conversations. 
Topic modeling methods~\cite{ptm} are appropriate. 
Therefore, inspired by ~\cite{Smith2018NeuralMF}, we employ a topic model based on variational auto-encoder (VAE)~\cite{kingma2013auto} and make adjustments according to our task to conduct the unsupervised global semantic modeling.
Compared to the traditional methods such as Latent Dirichlet Allocation (LDA) ~\cite{lda}, VAE-based topic model is less time-consuming for training and more flexible for inferring latent representations for new documents.

As shown in the upper side of Figure~\ref{fig:overview_of_our_model},  the input of VAE-based topic model is a document presentation $v$ and the output ${v'}$ is the reconstruction of input.
We regard each conversation as a document and represent it by $tf$-$idf$ features. 
The encoding process can be formalized as:
\begin{equation}
\label{eq:ntm}
\setlength{\abovedisplayskip}{3pt}
\setlength{\belowdisplayskip}{3pt}
\begin{aligned}
&h_v = f_h(v), \\
&\mu = f_\mu(h_v), \ \ \log(\sigma^2) = f_\sigma(h_v), \\
&z  = \mu + \sigma \cdot \epsilon, \ \ \epsilon \sim \mathcal{N}(0,1),\\
\end{aligned}
\end{equation}
where $f_*(\cdot)$ denotes non-linear transformation, $\mu$ and $\sigma$ are mean and standard deviation vectors of multivariate normal distribution respectively, $z$ is a latent vector sampled from the multivariate normal distribution by reparameterization trick.
We use the latent vector $z$ to reconstruct the document:
\begin{equation}
\label{eq:ntm2}
\setlength{\abovedisplayskip}{3pt}
\setlength{\belowdisplayskip}{3pt}
\begin{aligned}
h_v' &= f_{h'}(z), \\
v' &= f_{v'}(h_v'). \\
\end{aligned}
\end{equation}

We learn all parameters via optimizing the evidence lower bound (ELBO)~\cite{kingma2013auto}.
After the training, we draw a word-topic weight matrix $\mathbf{W} \in \mathbb{R}^{K\times |V'|}$ from the output layer $f_{v'}$. The  matrix represents the topical saliency for each word, where $K$ is the number of topics, $V'$ is the vocabulary of topic model and $|V'|$ is the vocabulary size. Each column $u \in \mathbb{R}^{K}$ in $\mathbf{W}$ can be regarded as a topic-based representation for the corresponding word.

Given topic-relevant word representations, we extend words for every dialogue.
After removing stop-words in the dialogue, we filter a vocabulary set $V^P \subset V'$ from predefined persona descriptions.
For each word $w \in V^P$, we select most relevant $m$ external words based on cosine similarities of topic-relevant word representations.
Then, we re-rank all external persona words of $V^P$ according to the cosine similarity score. If an external word is selected more than once, we just record the highest score. Thereafter, we select the top $n_w$ words of them. 
Finally, we convert each extended persona words into key-value representation by two multi-layer perceptron neural networks and store these representations in an \textbf{external persona words memory} $\mathbf{M^e}$.

\subsection{Persona Exploitation}
Given predefined persona descriptions and extended persona-relevant words, persona exploitation aims to integrate them to generate informative responses.
In this section, we detail three components of persona exploitation stage: multi-source sequence encoder, persona information retrieval and persona-oriented response decoder.

\paragraph{Multi-Source Sequence Encoder.}
The input contains persona descriptions and historical utterances, we design two independent encoders for them.

\textit{\textbf{Persona memory encoder.}} 
We encode the predefined persona information into sentence and word granularity presentations and store them in two memories respectively.
For each sentence $P_i$, we obtain a sentence representation $e_{P_i}$ by a bidirectional Gated Recurrent Networks (Bi-GRU)~\cite{DBLP:journals/corr/ChungGCB14}. Then we convert $e_{P_i}$ into key $m_i^{S}$ and value $c_i^{S}$ by two multi-layer perceptron neural networks, and store them in the sentence granularity persona memory $\mathbf{M^s}$.
Simultaneously, for word $p^i_j$, we obtain word representation $e_{p^i_j}$ from the $j$-th step of Bi-GRU for the $i$-th sentence. Same as above, we convert each word representation into key and value, and store them in the word granularity persona memory $\mathbf{M^w}$.

\textit{\textbf{Historical utterances encoder.}} In order to capture the relationship among the historical utterances $X$, we use a hierarchical recurrent encoder~\cite{DBLP:journals/corr/SerbanSBCP15} to conduct the semantic modeling.
From the second level of the hierarchical Bi-GRU, we obtain the final representations $e_{X}$ for the whole historical utterances and a sentence vector $C_i$ for each utterance $X_i$.

\paragraph{Persona Information Retrieval.}
\begin{figure}[!t]
  \includegraphics[width=0.9\linewidth]{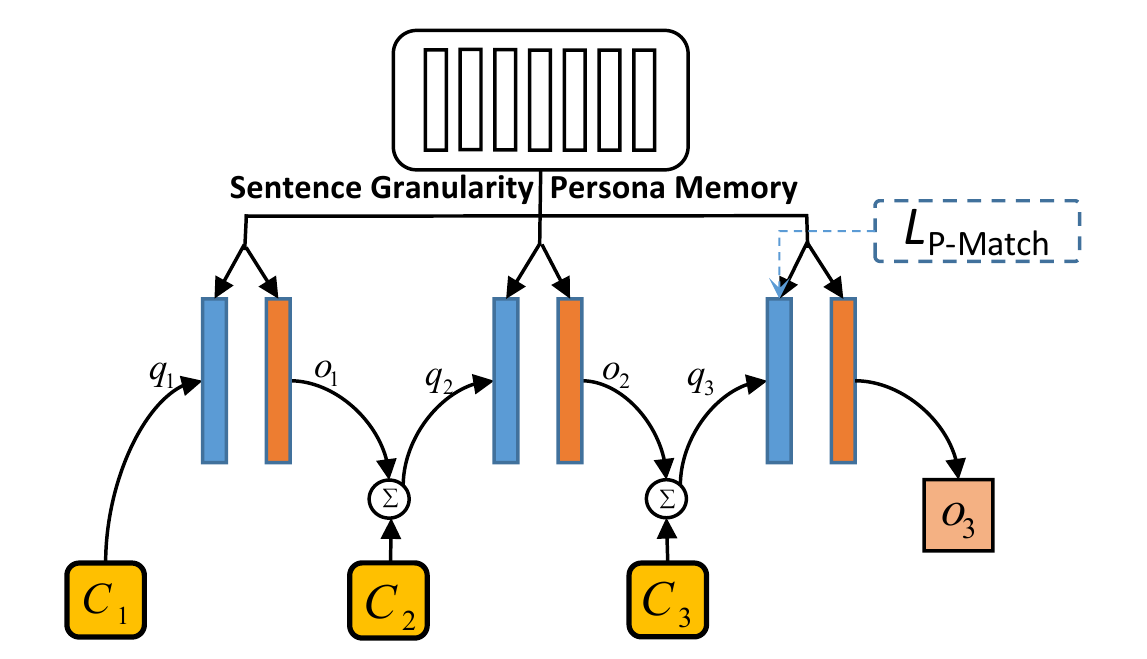}
  \caption{An overview of Persona Information Retrieval.} 
  \label{fig:example_PIR}
\end{figure}
After obtaining the representations of historical utterances and sentence granularity persona memory via the previous component, we use historical utterances to select persona information for response.
Considering the key-value memory retrieval is a frequent component in the following modules, we provide the general definition here. Assume that query vector is $q$ and memory $\mathbf{M}$ contains key $m$ and value $c$, the retrieval operation \textbf{retri} $(q, \mathbf{M}) = o$ is defined as:
\begin{equation}
\begin{aligned}
o & = \sum\nolimits_{\substack{i}} a_i c_i, \\
a_i & = \frac{\exp{(s_i)}}{\sum\nolimits_{j} \exp{(s_j)}}, \\
s_j & = q^T m_j,
\end{aligned}
\label{eq:retri}
\end{equation}
where the output vector $o$ is a weighted sum of values in memory and represents retrieved information.

As shown in Figure~\ref{fig:example_PIR}, we use each historical utterances to retrieve user's persona information in turn.
During the chat process, some of persona information used in the history has an impact on the choice of persona information for response.
In order to take advantage of this impact, in the $i$-th step, we combine historical utterance presentation $C_i$ and result of previous retrieval step $o_{i-1}$ as query vector $q_i$, which can be formalized as:
\begin{equation}
\setlength{\abovedisplayskip}{3pt}
\setlength{\belowdisplayskip}{3pt}
q_i=\left\{
\begin{array}{rcl}
& C_i, & {i = 1};\\
& C_i + o_{i-1}, & {i > 1}.
\end{array} \right.
\label{eq:query_vextor1}
\end{equation}
Then we retrieve the sentence granularity persona memory $\mathbf{M^s}$ by query vector $q_i$:
\begin{equation}
\setlength{\abovedisplayskip}{3pt}
\setlength{\belowdisplayskip}{3pt}
o_i = retri(q_i,\mathbf{M^s} ).
\label{eq:retri_PS}
\end{equation}

Finally, we concatenate the result of last retrieval step $o_k$ and the whole historical utterances representations $e_X$: 
\begin{equation}
\setlength{\abovedisplayskip}{3pt}
\setlength{\belowdisplayskip}{3pt}
s_0 = [e_X; o_k],
\end{equation}
where $s_0$ is a merged vector used as the initial state of decoder.

\paragraph{Persona-Oriented Response Decoder.}
The decoder is a GRU based sequence prediction framework with an attention mechanism on the historical utterances and a mutual-reinforcement multi-hop memory retrieval mechanism.
Given the current input $y_{t-1}$ as well as the previous hidden state $s_{t-1}$, the recurrent calculation of GRU is defined as:
\begin{equation}
\setlength{\abovedisplayskip}{3pt}
\setlength{\belowdisplayskip}{3pt}
s_t = GRU(y_{t-1},s_{t-1}).
\end{equation}
Then we design an attention mechanism to absorb relevant information from historical utterances and a mutual-reinforcement multi-hop memory retrieval mechanism to obtain the relevant persona information from predefined and explored external persona information.

\textit{\textbf{Attention on the historical utterances.}} produces a historical utterances vector $u^X$ at each decoding step by attending to historical utterances. We formalize it as:
\begin{equation}
\setlength{\abovedisplayskip}{3pt}
\setlength{\belowdisplayskip}{3pt}
\label{eq:atten}
\begin{aligned}
u^X & = \sum\limits_{i=1}^{n} a_i h^X_i, \\
a_i & = \frac{\exp{(s_i)}}{\sum\nolimits_{j=1}^{n} \exp{(s_j)}}, \\
s_j & = v^T \tanh{(W_s s_t + W_t h^X_j + b)},
\end{aligned}
\end{equation}
where $h^X_j (j = 1,2,\dots,n)$ is the $j$-th word hidden state of historical utterances obtained from the first level of the hierarchical encoder for $X$.
 
\textit{\textbf{Mutual-reinforcement multi-hop memory retrieval.}} Recall that we build an external persona words memory $\mathbf{M^e}$ for persona exploration and a word granularity persona memory $\mathbf{M^w}$ in encoder.
There is an association between the two memories.
For example, if we retrieve a word in the predefined persona descriptions which is related to current conversation, the information in external persona memory related to this word will be more likely to be applied, vice versa. 
Therefore, the results of the two types of persona information retrievals are mutually influential and we propose a mutual-reinforcement multi-hop memory retrieval mechanism to model this influence.

First, we use the current hidden state $s_t$ as query vector $q$ to retrieve $\mathbf{M^w}$ and $\mathbf{M^e}$ respectively:
\begin{equation}
\setlength{\abovedisplayskip}{3pt}
\setlength{\belowdisplayskip}{3pt}
\begin{aligned}
o^{w} & = retri(q,\mathbf{M^w}), \\
o^{e} & = retri(q,\mathbf{M^e}).
\end{aligned}
\label{eq:retri_cross}
\end{equation}
Considering that the result of one memory retrieval (e.g. $o^{w}$) will affect the next retrieval of another memory (e.g. $\mathbf{M^e}$), we update query vector by adding the two retrieved results $o^{w}$ and $o^{e}$:
\begin{equation}
\setlength{\abovedisplayskip}{3pt}
\setlength{\belowdisplayskip}{3pt}
q^{new} = q^{old} + o^{w} + o^{e}.
\label{eq:retri_cross_update}
\end{equation}
This update means that the results of the two retrievals will affect each other in the next hop. 
In our experiment, we use three hops unless otherwise stated.

Finally, based on the exploitation of predefined and extended persona information, the output word distribution $p_{y_t}$ at time step $t$ of the decoder is produced by:
\begin{equation}
\begin{aligned}
&\tilde{s}_t = f_o([s_t; u^X; o^{w}; o^{e}]), \\
&p_{y_t} = softmax(\tilde{s}_t),
\end{aligned}
\end{equation}
where $f_o$ is the neural non-linear operation on the output layer.


\subsection{Persona-Orientated Loss.}
In order to impel the model to exploit the persona information more precisely, besides the Negative Log-Likelihood loss (NLL), we propose two new persona-oriented loss functions: Persona-oriented Matching loss (P-Match) and Persona-oriented Bag-of-Words loss (P-BoWs).
P-Match loss supervises the choice of predefined persona sentences in persona information retrieval module and P-BoWs loss supervises to generate more persona-related words in decoder.

\paragraph{P-Match Loss.}
Recall that in the persona information retrieval module (Eq.~\ref{eq:retri_PS}), 
we can get a match weight over the sentence granularity persona memory $\mathbf{M^s}$ in every step. Assume that the match weight in the last step is $a^s \in \mathbb{R}^{|P|}$.
Intuitively, if the ground truth response contains the information from persona sentence $P_i$, then $a_i^s$ should obtain a large value. Is it possible to employ the relation between the ground truth response and the persona sentences to improve the modeling of persona information retrieval? To tackle this, we design the persona-oriented matching loss (P-Match). 
The $0$-$1$ label $a \in \mathbb{R}^{|P|}$ is decided based on a threshold $\theta_a$ of the similarity between the persona sentences and the ground truth response. Jaccard Index\footnote{http://en.wikipedia.org/wiki/Jaccard\_index} is employed for the similarity calculation.
The P-Match loss is defined as:
\begin{equation}
\setlength{\abovedisplayskip}{3pt}
\setlength{\belowdisplayskip}{3pt}
\mathcal{L}_{P-Match} = - \sum\limits_{i=1}^{|P|} a_i \log{a_i^s}.
\end{equation}

\paragraph{P-BoWs Loss.}
Inspired by~\cite{DBLP:journals/corr/abs-1805-04871}, we design a persona-oriented Bag-of-Words loss function to enhance the ability of persona information capturing. Specifically,
We label each response with a vocabulary-size vector $b \in \mathbb{R}^{|V|}$, where the non-stop words in the current response will get values $1$. If words are persona-based information, we increase the weight to $1+\lambda$, where $\lambda$ is a positive value. We use a multi-label classifier to generate BoWs representation $P_b$ (sentence-level probability) by summing the scores of all positions of the generated sentence in decoder:
$ p_b = sigmoid(\sum\limits_{t=1}^{|Y|} \tilde{s}_t) $.
We define P-BoWs loss using cross entropy:
\begin{equation}
\setlength{\abovedisplayskip}{3pt}
\setlength{\belowdisplayskip}{3pt}
\begin{aligned}
\mathcal{L}_{P-BoWs} = - \frac{1}{|V|} \sum\limits_{i=1}^{|V|} \lbrack b_i\log{p_{b_i}} + \\ 
(1-b_i)\log{(1-p_{b_i})} \rbrack.
\end{aligned}
\end{equation}

\subsection{Joint Training}
Negative log-likelihood loss (NLL) is employed as the basic optimization objective:
\begin{equation}
\setlength{\abovedisplayskip}{3pt}
\setlength{\belowdisplayskip}{3pt}
\mathcal{L}_{NLL} = - \frac{1}{|Y|} \sum\limits_{t=1}^{|Y|} y_t \log{p_{y_t}}. 
\end{equation}

Finally, a unified optimization objective is designed by integrating P-Match loss, P-BoWs loss and the NLL loss:
\begin{equation}
\setlength{\abovedisplayskip}{3pt}
\setlength{\belowdisplayskip}{3pt}
\mathcal{L} = \mathcal{L}_{NLL} + \gamma_1 \mathcal{L}_{P-Match} + \gamma_2 \mathcal{L}_{P-BoWs},
\end{equation}
where $\gamma_1$ and $\gamma_2$ are trade-off parameters controlling the balance between three loss functions.

\section{Experiments}
In this section, we first introduce two datasets used in our experiment and list setups and baseline models.
Next, we evaluate the performance of various models by automated evaluation and human evaluation.

\begin{table*}
\centering
\caption{Automatic evaluation results. The best results are bold.}
\resizebox{0.8\linewidth}{!}{\begin{tabular}{ccccccccc}
    \toprule
        \textbf{Model}&\textbf{BLEU1}&\textbf{BLEU2}&\textbf{BLEU3}&\textbf{BLEU4}&\textbf{F1}&\textbf{Average}&\textbf{Extrema}&\textbf{Greedy}\\
        \midrule
        Seq2Seq&20.1381&9.9395&5.2887&2.9840&17.7972&0.8551&0.4980&0.6751\\
        HRED&19.0920&9.5668&5.0191&2.7779&17.9184&0.8531&0.4882&0.6714\\
        Profile Memory&20.8713& 9.8526&4.9942&2.6852&17.1553&0.8675&0.4835&0.6752\\
        Per.-CVAE&17.2315& 7.2602&3.2081&1.4541&14.6121&0.8458&0.4688&0.6516\\
        \midrule
        PED&21.4611&10.6992&5.7845&3.3344&\textbf{18.4759}&0.8593&0.4993&0.6838\\
        PED+PE&21.8970&10.9987&5.9965&3.5334&18.4140&0.8643&0.4999&0.6856\\
        PED+PE+P-BoWs&21.9768&11.0710&6.0154&\textbf{3.5574}&18.2781&0.8626&0.4986&0.6822\\
        PED+PE+P-Match&22.4668&11.2560&5.9846&3.3031&18.2615&0.8592&0.4940&0.6803\\
        PEE&\textbf{23.1926}& \textbf{11.5166}&\textbf{6.1248}&3.4977&18.4130&\textbf{0.8691}&\textbf{0.5010}&\textbf{0.6906}\\
        \bottomrule
    \end{tabular}}
\label{tab:automated_evaluation}
\end{table*}

\begin{table}
    \small
    \caption{Human evaluation on four aspects: Fluency, Engagingness, Consistency and Persona Detection (PD). The value in parentheses is standard deviation.}
    \centering
\resizebox{0.95\columnwidth}{!}{\begin{tabular}{ccccc}
        \toprule
        \textbf{Model}& \textbf{Fluency}&\textbf{Engagingness}&\textbf{Consistency}&\textbf{PD(\%)}\\
        \midrule
        Seq2Seq& 4.08(0.71)& 3.02(0.96)&3.00(1.03)&52.94(0.32)\\
        HRED&3.96(0.71)& 2.73(1.05)&2.60(1.16)&64.71(0.32)\\
        Profile Memory& 4.04(0.68)&3.08(1.01)&3.10(1.10)&58.82(0.40)\\
        Per.-CVAE& 3.61(1.02)&2.63(1.09)&2.78(1.29)&\textbf{85.29(0.34)}\\
        PEE& \textbf{4.13(0.76)}&\textbf{3.46(1.07)}&\textbf{3.44(1.13)}&76.47(0.36)\\
        \bottomrule
    \end{tabular}}
    \label{tab:human_evaluation}
\end{table}

\begin{table}
    \small
    \centering
    \caption{Automatic evaluation results of PEE with different hops in mutual-reinforcement multi-hop retrieval mechanism.}
\resizebox{0.95\columnwidth}{!}{\begin{tabular}{ccccccccc}
        \toprule
        \textbf{Hops}&\textbf{BLEU1}&\textbf{BLEU2}&\textbf{BLEU3}&\textbf{BLEU4}&\textbf{F1}&\textbf{Average}&\textbf{Extrema}&\textbf{Greedy}\\
        \midrule
        PEE-1&22.5956&11.2877&6.0405&3.4315&18.32&0.8631&0.5009&0.6902\\
        PEE-2&22.9758&11.4999&\textbf{6.2383}&\textbf{3.6327}&\textbf{18.68}&0.8654&0.4979&0.6861\\
        PEE-3&\textbf{23.1926}&\textbf{11.5166}&6.1248&3.4977&18.41&\textbf{0.8691}&\textbf{0.5010}&\textbf{0.6906}\\
        PEE-4&22.3422&11.0628&5.8804&3.3678&18.25&0.8618&0.4985&0.6824\\
        PEE-5&22.2892&11.1789&5.9878&3.4148&18.55&0.8591&0.4993&0.6811\\
        \bottomrule
    \end{tabular}}
    \label{tab:hop}
\end{table}

\subsection{Datasets}
Our experiments use two public multi-turn dialogue datasets: Persona-Chat\footnote{https://github.com/facebookresearch/ParlAI/tree/ \\ master/projects/personachat} ~\cite{P18-1205} and DailyDialog\footnote{http://yanran.li/dailydialog}  ~\cite{Li2017DailyDialogAM}. 
The Persona-Chat dataset contains 10907 dialogues between pairs of speakers, where 968 dialogues are set aside for validation and 1000 for testing.
Each speaker is described by 3-5 persona sentences.
(e.g. ``\textit{I like reading.}'' or ``\textit{I am a nurse.}'', etc). 
The total number of personas is 1155, and 100 personas for validation and 100 for testing.
The DailyDialog dataset is constructed by raw data crawled from various websites, which serves for English learners to practice English dialog in daily life.
It contains 13,118 multi-turn dialogues without persona descriptions, the number of turns are roughly 8 and the average number of tokens of each utterance is about 15.

Our experiments are performed on the persona-chat dataset. In order to expand the knowledge space, we merge DailyDialog and the training set of Persona-Chat as the basic knowledge source to pretrain the topic model for persona exploration. 

\subsection{Baselines}
We consider the following comparison methods and their inputs consist of predefined persona descriptions, historical conversation utterances and the current query utterance.

\noindent \textbf{Seq2Seq}~\cite{bahdanau2014neural}: the standard Sequence-to-Sequence Model with attention.
We concatenate persona descriptions and historical utterances as a sequence input and generate the response.

\noindent\textbf{HRED}~\cite{DBLP:journals/corr/SerbanSBCP15}: Hierarchical Recurrent Encoder-Decoder model with attention. 
The input contains all sentences in persona and history conversation.

\noindent \textbf{Profile Memory}~\cite{P18-1205}: Generative Profile Memory network is a generative model that encodes each of persona descriptions as a individual memory representation in a memory network.

\noindent \textbf{Per.-CVAE}~\cite{Exploiting}: Persona-CVAE is a memory-augmented architecture which focus on the diverse generation of conversational responses based on chatbot’s persona. In our experiment, we sample one time from the latent z to generate a response.

\noindent \textbf{PED}: Persona-oriented Encoder-Decoder Model. i.e., our PEE framework without the persona exploration, the P-BoWs loss and the P-Match loss.
Without external persona words memory, mutual-reinforcement multi-hop memory retrieval mechanism is equivalent to normal multi-hop memory retrieval mechanism. 

\noindent \textbf{PED+PE}: Our PEE framework without the P-BoWs loss and P-Match loss.

\noindent \textbf{PED+PE+P-BoWs}: Our PEE framework without the P-Match loss.

\noindent \textbf{PED+PE+P-Match}: Our PEE framework without the P-BoWs loss.

\noindent \textbf{PEE}: PED + PE + P-BoWs + P-Match, i.e., our proposed PEE framework.

\subsection{Experimental Settings}
We treat each complete dialog (including personas) as a document, remove the stop words and select the top 10,000 frequent words to train the VAE-based topic model.
For the number of topics, we follow previous settings ~\cite{Yan:2013:BTM:2488388.2488514}, \cite{Smith2018NeuralMF} to set K = 50.
In our experiments, we use GloVe ~\cite{pennington2014glove} for word embedding and employ bi-directional GRU for encoders, and we set hidden states size is 512 and batch size is 64.
We use Adam optimizer ~\cite{journals/corr/KingmaB14} to train the model and set the learning rate is 0.0001.
For testing, we use beam search with beam size 2.
For all the other hyperparameters, we tune them on the development set by grid search.
The number of extended persona-related words for each dialogue $n_w$ is 100.
The additional weight $\lambda$ in the P-BoWs target is 1 and threshold $\theta_a$ in labeling P-Match target process is 0.03.
During training, trade-off parameter $\gamma_1$ is 0.1 and $\gamma_2$ is 0.1.

\begin{table*}[!t]
    \centering
    \small
    \caption{Case studies.}
\resizebox{0.88\linewidth}{!}{\begin{tabular}{c|l|l}
            \toprule
             & \textbf{Case 1} & \textbf{Case 2}\\
             \midrule
             \multirow{2}{*}{Personas for}  
              &I worked at hollister in high school.& I write short \textbf{stories} for a living.  \\
              &I am a professional skater.& I used to work at a carnival.  \\
            \multirow{2}{*}{Speaker B }  
              &I play bluegrass \textbf{music}.& I like to drink scotch to relax.  \\
              &I do not like chips. & I like dark superhero movies. \\
            \midrule

            \multirow{2}{*}{}
            & A: Hi. How is your night going? & A: Hi, how are you doing tonight? I am good so far.  \\
            & B: Good, just left a bluegrass concert. & B: I am good, relaxing with a glass of scotch to end the night.  \\
            \multirow{1}{*}{Historical}
            & A: Ooh. Interesting. What else do you do for fun? & A: That is nice and relaxing I love to get a good workout.  \\
            \multirow{1}{*}{}
            & B: Skate. I love it. What do you do? & B: Yes. I have been busy working on a new short story to release. \\
            \multirow{1}{*}{Utterances}
            & A: That is so exciting! I am currently a stay at home mom. & A: That sounds very interesting hope all is good for you.\\
            \multirow{2}{*}{}
             & B: cool, how many kids do you have? & B: Yeah. What kind of movies are you interested in? \\
            & A: I have three \textbf{kids} and pregnant with my fourth. I love being a \textbf{mom}.& A: I like to watch romance and some \textbf{scary movies} is okay for me.\\
            \midrule

            \multirow{1}{*}{Explored words}
            & concert, band, \textbf{piano}, guitar, rap ... & ebook, \textbf{thriller}, \textbf{horror}, creepy, comic ...  \\ 
            \midrule
            
            \multirow{4}{*}{Response}
            & seq2seq: That is cool. Do you have any pets? & seq2seq: I like movies too. I am a loner.  \\
            & HRED: That is cool. Do you have any pets? & HRED: That is cool. I am a fan of a movie.  \\
            & Profile Memory: I have a lot of kids, but i have a daughter that is so cool. & Profile Memory: I have been watching movies on TV. \\
            & Per.-CVAE: I am a coach I might play tomorrow. I do roofing. & Per.-CVAE: I like movies too, I am not sure I like that. What are your hobbies? \\
            & PEE: Wow that is a great thing. I like to play \textbf{piano} with my \textbf{family}.  & PEE: That is scary. I write a lot of \textbf{horror stories}. \\
            \bottomrule
        \end{tabular}}
    \label{tab:case_study}
\end{table*}

\subsection{Evaluation Metrics}
We use different evaluation metrics (automated and human) to demonstrate the effectiveness of our model. In this subsection, we will give a brief introduction to those metrics. 

\paragraph{Automatic Metrics.}
We report three different automatic metrics:

\noindent\textbf{BLEU@N}: BLEU is an algorithm which has been widely used in machine translation and dialogue system to evaluate the quality of the generated text. It measures the N-gram overlap between the generated response and ground truth. 

\noindent\textbf{F1-Measure}: It measures the accuracy of the generated response considering both the precision and the recall. We treat the predicted and target response as bags of tokens, and compute their F1 score. 
  
\noindent\textbf{Embedding-based similarity}: Embedding Average (Average), Embedding Extrema (Extrema), and Embedding Greedy (Greedy)~\cite{DBLP:journals/corr/LiuLSNCP16}. These embedding-based metrics measure semantic similarity between the generated response and the ground truth.

\paragraph{Human Metrics.}
It is not enough to only automatically evaluate dialogue systems, so we randomly sample about 100 dialogues from test data and hire 5 volunteers to evaluate.
We use four metrics: fluency, engagingness, consistency, and persona detection. 

\noindent\textbf{Fluency}: It measures the quality of generated sentence, e.g., whether grammar is correct. 

\noindent\textbf{Engagingness}: It measures whether the generated sentence is appropriate and interesting. 

\noindent\textbf{Consistency}: It measures whether the generated sentence has some relationships with the history and persona description.

\noindent\textbf{Persona detection}: For each dialogue, given generated responses and two set of persona sentences (one is real and another is fake), we ask students to choose which one is the real description of chatbot.

The first three metrics are scored between 1-5.
For persona detection, Scoring 1 means the choice is correct and 0 means the choice is wrong and 0.5 means people can not judge.

\section{Results and Analysis}

\subsection{Experimental Results and Ablation Study}
\paragraph{Automatic evaluation.}
Comparative automatic evaluation results are presented in Table \ref{tab:automated_evaluation}. 
Our model outperforms baselines on all automatic metrics. This demonstrates that our model generates more appropriate responses by persona exploration and exploitation. Especially, our model improves approximately 15.17\% over seq2seq on BLEU1. 
Comparing with PED, PED+PE has better scores on most metrics. This is because explored persona information contributes to generate more informative responses. 
Comparing with PED+PE, both PED+PE+p-BoWs and PED+PE+P-Match perform better, which is because P-BoWs loss and P-Match loss supervise model to exploit persona information more precisely.

According to automatic evaluation results of PEE with different hops in mutual-reinforcement multi-hop retrieval mechanism in Table\ref{tab:hop}, PEE-2 outperforms PEE-1 on most metrics. This demonstrates the interaction between two types persona information improves the performance of our model. 
Analyzing various indicators, the PEE works best when hop is 3. 
When the number of hops exceeds 3, the effect drops, that may because the query vector contains little information of current decoder state $s_t$ after several update operations.

\paragraph{Human evaluation.}
 The results of human evaluation are listed in Table \ref{tab:human_evaluation}. Our model significantly outperforms most of the baselines in terms of all the metrics. Particularly, our model increases approximately 30.01\% over profile memory on persona detection. This demonstrates that persona exploration and exploitation is beneficial to improve the usage of persona information and enrich the responses. Per.-CVAE has the highest persona detection metric, but it pays too much attention to persona, resulting in very poor grammar, relevance, and fluency of the generated responses.

\subsection{Persona Analysis}

\begin{figure}[!t]
    \centering
    \includegraphics[width=1\linewidth]{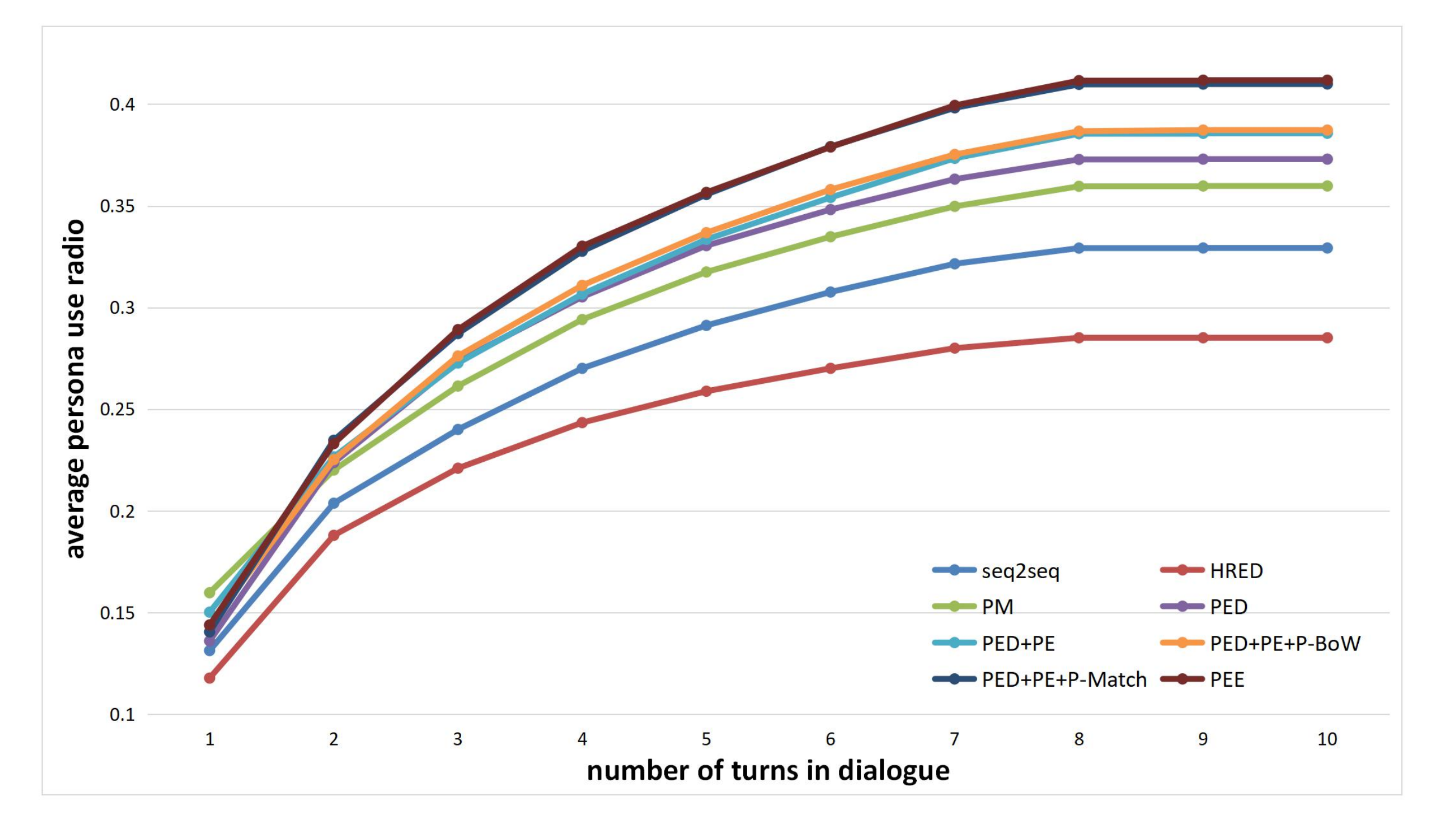}
    \caption{Average persona use ratio of all models in different turns.}
    \label{fig:pur}
\end{figure}
In order to further evaluate the ability of the model to combine persona, for each multi-turn dialogue, we count the number of words that appear in both persona sentences and generated responses, and divide the number by the total number of words in persona sentence to get \textbf{Persona use ratio}. 
It measures the probability of persona words being used and punishes the repeated use of the same persona information in responses of different turns.
We calculate the average persona use ratio of all models with different turns and show them in Figure \ref{fig:pur}. 
Per.-CVAE pays too much attention to persona, which seriously affects the quality of the generated responses. We do not consider Per.-CVAE here.
We can see that our model outperforms all the baseline methods. There are three reasons: first, persona information retrieval module thinks about the influence of persona in history when selecting persona information; second, external persona words contribute to utilize persona description that has an indirect relationship with current topic. third, the P-BoWs loss and the P-Match loss encourage the model to generate more persona-related words.

\subsection{Case Study}
\begin{figure}[!t]
    \centering
    \includegraphics[width=1\linewidth]{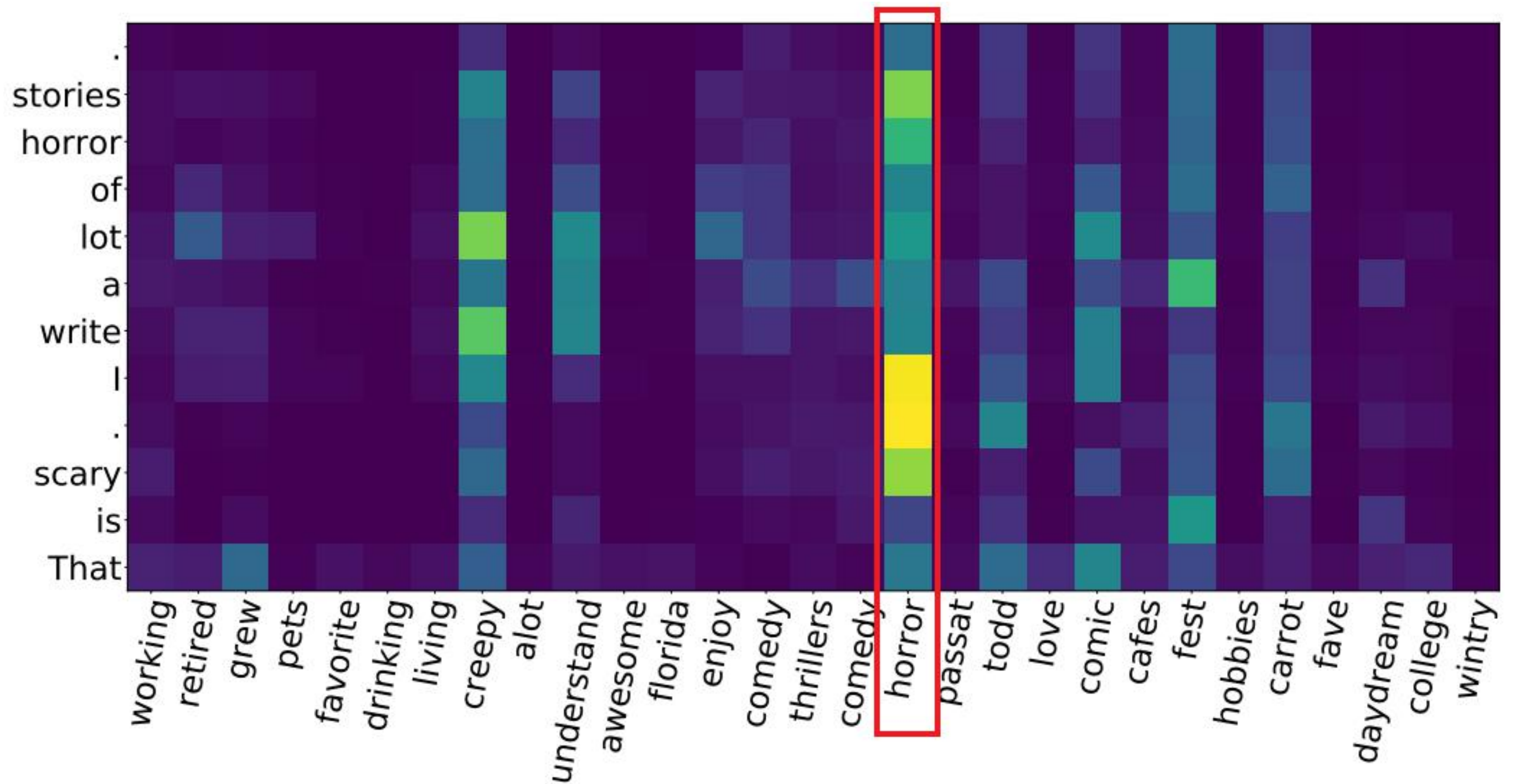}
    \caption{Visualization of matching weights on external persona words memory in the last step of mutual-reinforcement multi-hop memory retrieval.}
    \label{Fig.visualization}
\end{figure}

Table \ref{tab:case_study} depicts some cases generated by PEE, Seq2Seq, HRED, and Profile Memory. From the comparisons, we can see that PEE model can use explored persona information to generate more persona-oriented informative responses. 
For example, in case 1, one of persona descriptions for speaker B is ``I play bluegrass music.'', speaker A mentioned ``kids'' and ``mom'' in the query utterance. The explored persona word ``piano'' is related to ``music'' and ``family'', the word ``family'' has correlation with ``kids'' and ``mom''. So, the response generated by PEE follows the clues above and implies persona information simultaneously. 
What's more, it leads the topic to a new field that speakers are familiar with, making the next reply have more content to facilitate. When the previous topic "work" is drawing to an end, our model can use persona and its extended words to convert topics to "music", however, responses of other baseline models do not reflect personality information.

Similarly, in case 2, explored persona word ``horror'' is related with ``scary movies'' and ``stories'', so our PEE model uses word ``horror'' to rich response.
In order to show the contribution of explored persona words more clearly and directly, we show the matching weights on external persona words memory in the last step of mutual-reinforcement multi-hop memory retrieval mechanism in Figure~\ref{Fig.visualization}.

\section{CONCLUSION}
In this work, we propose a neural topical expansion framework, namely Persona Exploration and Exploitation (PEE), for the unstructured persona-oriented dialogue systems.
Different from previous work that trained purely on predefined persona description, our model extends external persona information by a VAE-based topic model. 
By fusing predefined persona descriptions and external persona information, the responses our model generates can more accurately and properly represent the user persona while maintaining the consistency of the dialogue.
Experimental comparisons and analysis demonstrated that our approach outperforms a set of state-of-the-art baselines in terms of both automated metrics and human evaluations.
For future work, we will extend persona information dynamically and jointly train persona exploration and exploitation.

\section{ACKNOWLEDGEMENT}
This work is supported by the Natural Science Foundation of China (61972234, 61902219, 61672324, 61672322), the Tencent AI Lab Rhino-Bird Focused Research Program (JR201932), the Foundation of State Key Laboratory of Congitive Intelligence, iFLYTEK, P.R.China (COGOSC- 20190003), the Fundamental Research Funds of Shandong University, Ahold Delhaize, the Association of Universities in the Netherlands (VSNU), and the Innovation Center for Artificial Intelligence (ICAI).
\bibliography{ecai}

\begin{thebibliography}{33}
\providecommand{\natexlab}[1]{#1}
\providecommand{\url}[1]{\texttt{#1}}
\expandafter\ifx\csname urlstyle\endcsname\relax
  \providecommand{\doi}[1]{doi: #1}\else
  \providecommand{\doi}{doi: \begingroup \urlstyle{rm}\Url}\fi

\bibitem[Bahdanau et~al.(2014)Bahdanau, Cho, and Bengio]{bahdanau2014neural}
D.~Bahdanau, K.~Cho, and Y.~Bengio.
\newblock Neural machine translation by jointly learning to align and
  translate.
\newblock \emph{arXiv preprint arXiv:1409.0473}, 2014.

\bibitem[Blei et~al.(2002)Blei, Ng, and Jordan]{lda}
D.~M. Blei, A.~Y. Ng, and M.~I. Jordan.
\newblock Latent dirichlet allocation.
\newblock In \emph{Advances in Neural Information Processing Systems 14}, pages
  601--608, 2002.

\bibitem[Chu et~al.(2018)Chu, Vijayaraghavan, and
  Roy]{DBLP:journals/corr/abs-1810-08717}
E.~Chu, P.~Vijayaraghavan, and D.~Roy.
\newblock Learning personas from dialogue with attentive memory networks.
\newblock \emph{arXiv preprint arXiv:1810.08717}, 2018.

\bibitem[Chung et~al.(2014)Chung, G{\"{u}}l{\c{c}}ehre, Cho, and
  Bengio]{DBLP:journals/corr/ChungGCB14}
J.~Chung, {\c{C}}.~G{\"{u}}l{\c{c}}ehre, K.~Cho, and Y.~Bengio.
\newblock Empirical evaluation of gated recurrent neural networks on sequence
  modeling.
\newblock \emph{arXiv preprint arXiv:/1412.3555}, 2014.

\bibitem[Joshi et~al.(2017)Joshi, Mi, and
  Faltings]{DBLP:journals/corr/JoshiMF17}
C.~K. Joshi, F.~Mi, and B.~Faltings.
\newblock Personalization in goal-oriented dialog.
\newblock \emph{arXiv preprint arXiv:1706.07503}, 2017.

\bibitem[Kingma and Ba(2014)]{journals/corr/KingmaB14}
D.~P. Kingma and J.~Ba.
\newblock Adam: A method for stochastic optimization.
\newblock \emph{arXiv preprint arXiv:/1412.6980}, 2014.

\bibitem[Kingma and Welling(2013)]{kingma2013auto}
D.~P. Kingma and M.~Welling.
\newblock Auto-encoding variational bayes.
\newblock \emph{arXiv preprint arXiv:1312.6114}, 2013.

\bibitem[Li et~al.(2016)Li, Galley, Brockett, Spithourakis, Gao, and Dolan]{li}
J.~Li, M.~Galley, C.~Brockett, G.~Spithourakis, J.~Gao, and B.~Dolan.
\newblock A persona-based neural conversation model.
\newblock In \emph{Proceedings of the 54th Annual Meeting of the Association
  for Computational Linguistics (Volume 1: Long Papers)}, pages 994--1003,
  2016.

\bibitem[Li et~al.(2017)Li, Su, Shen, Li, Cao, and Niu]{Li2017DailyDialogAM}
Y.~Li, H.~Su, X.~Shen, W.~Li, Z.~Cao, and S.~Niu.
\newblock Dailydialog: A manually labelled multi-turn dialogue dataset.
\newblock In \emph{Proceedings of the 8th International Joint Conference on
  Natural Language Processing}, 2017.

\bibitem[Lin et~al.(2019)Lin, Madotto, Wu, and Fung]{lin2019personalizing}
Z.~Lin, A.~Madotto, C.-S. Wu, and P.~Fung.
\newblock Personalizing dialogue agents via meta-learningf.
\newblock In \emph{Proceedings of the 57th Annual Meeting of the Association
  for Computational Linguistics}, pages 5454--5459, 2019.

\bibitem[Liu et~al.(2016)Liu, Lowe, Serban, Noseworthy, Charlin, and
  Pineau]{DBLP:journals/corr/LiuLSNCP16}
C.~Liu, R.~Lowe, I.~V. Serban, M.~Noseworthy, L.~Charlin, and J.~Pineau.
\newblock How {NOT} to evaluate your dialogue system: An empirical study of
  unsupervised evaluation metrics for dialogue response generation.
\newblock \emph{arXiv preprint arXiv:/1603.08023}, 2016.

\bibitem[Luo et~al.(2018)Luo, Huang, Zeng, Nie, and
  Sun]{DBLP:journals/corr/abs-1811-04604}
L.~Luo, W.~Huang, Q.~Zeng, Z.~Nie, and X.~Sun.
\newblock Learning personalized end-to-end goal-oriented dialog.
\newblock \emph{arXiv preprint arXiv:/1811.04604}, 2018.

\bibitem[Ma et~al.(2018)Ma, Sun, Wang, and
  Lin]{DBLP:journals/corr/abs-1805-04871}
S.~Ma, X.~Sun, Y.~Wang, and J.~Lin.
\newblock Bag-of-words as target for neural machine translation.
\newblock \emph{arXiv preprint arXiv:/1805.04871}, 2018.

\bibitem[Mazar{\'{e}} et~al.(2018)Mazar{\'{e}}, Humeau, Raison, and
  Bordes]{DBLP:journals/corr/abs-1809-01984}
P.~Mazar{\'{e}}, S.~Humeau, M.~Raison, and A.~Bordes.
\newblock Training millions of personalized dialogue agents.
\newblock \emph{arXiv preprint arXiv:/1809.01984}, 2018.

\bibitem[Meng et~al.(2020)Meng, Ren, Chen, Monz, Ma, and de~Rijke]{refnet}
C.~Meng, P.~Ren, Z.~Chen, C.~Monz, J.~Ma, and M.~de~Rijke.
\newblock Refnet: A reference-aware network for background based conversation.
\newblock In \emph{Proceedings of the 34th AAAI Conference on Artificial
  Intelligence}, 2020.

\bibitem[Mo et~al.(2018)Mo, Zhang, Li, Li, and Yang]{Mo2018PersonalizingAD}
K.~Mo, Y.~Zhang, S.~Li, J.~Li, and Q.~Yang.
\newblock Personalizing a dialogue system with transfer reinforcement learning.
\newblock In \emph{Proceedings of the 32th AAAI Conference on Artificial
  Intelligence}, pages 5317--5324, 2018.

\bibitem[Olabiyi et~al.(2019)Olabiyi, Khazane, Salimov, and
  Mueller]{DBLP:journals/corr/abs-1905-01992}
O.~Olabiyi, A.~Khazane, A.~Salimov, and E.~T. Mueller.
\newblock An adversarial learning framework for {A} persona-based multi-turn
  dialogue model.
\newblock \emph{arXiv preprint arXiv:/1905.01992}, 2019.

\bibitem[Pennington et~al.(2014)Pennington, Socher, and
  Manning]{pennington2014glove}
J.~Pennington, R.~Socher, and C.~Manning.
\newblock Glove: Global vectors for word representation.
\newblock In \emph{Proceedings of the 2014 conference on empirical methods in
  natural language processing}, pages 1532--1543, 2014.

\bibitem[Qian et~al.(2018)Qian, Huang, Zhao, Xu, and Zhu]{ijcai2018-595}
Q.~Qian, M.~Huang, H.~Zhao, J.~Xu, and X.~Zhu.
\newblock Assigning personality/profile to a chatting machine for coherent
  conversation generation.
\newblock In \emph{Proceedings of the 27th International Joint Conference on
  Artificial Intelligence}, pages 4279--4285, 2018.

\bibitem[Ren et~al.(2020)Ren, Chen, Monz, Ma, and de~Rijke]{ThinkingGlobally}
P.~Ren, Z.~Chen, C.~Monz, J.~Ma, and M.~de~Rijke.
\newblock Thinking globally, acting locally: Distantly supervised
  global-to-local knowledge selection for background based conversation.
\newblock In \emph{Proceedings of the 34th AAAI Conference on Artificial
  Intelligence}, 2020.

\bibitem[Serban et~al.(2015)Serban, Sordoni, Bengio, Courville, and
  Pineau]{DBLP:journals/corr/SerbanSBCP15}
I.~V. Serban, A.~Sordoni, Y.~Bengio, A.~C. Courville, and J.~Pineau.
\newblock Hierarchical neural network generative models for movie dialogues.
\newblock \emph{arXiv preprint arXiv:/1507.04808}, 2015.

\bibitem[Serban et~al.(2016)Serban, Sordoni, Lowe, Charlin, Pineau, Courville,
  and Bengio]{serban2016hierarchical}
I.~V. Serban, A.~Sordoni, R.~Lowe, L.~Charlin, J.~Pineau, A.~C. Courville, and
  Y.~Bengio.
\newblock A hierarchical latent variable encoder-decoder model for generating
  dialogues.
\newblock \emph{arXiv preprint arXiv:/1605.06069}, 2016.

\bibitem[Smith et~al.(2018)Smith, Card, and Tan]{Smith2018NeuralMF}
N.~A. Smith, D.~Card, and C.~Tan.
\newblock Neural models for documents with metadata.
\newblock In \emph{Proceedings of the 56th Annual Meeting of the Association
  for Computational Linguistics}, 2018.

\bibitem[Song et~al.(2019)Song, Zhang, Cui, Wang, and Liu]{Exploiting}
H.~Song, W.~Zhang, Y.~Cui, D.~Wang, and T.~Liu.
\newblock Exploiting persona information for diverse generation of
  conversational responses.
\newblock In \emph{Proceedings of the the 28th International Joint Conference
  on Artificial Intelligence}, pages 5190--5196, 2019.

\bibitem[Steyvers and Griffiths(2006)]{ptm}
M.~Steyvers and T.~Griffiths.
\newblock Probabilistic topic models.
\newblock In T.~Landauer, D.~McNamara, S.~Dennis, and W.~Kintsch, editors,
  \emph{Latent Semantic Analysis: A Road to Meaning.}, 2006.

\bibitem[Urbanek et~al.(2019)Urbanek, Fan, Karamcheti, Jain, Humeau, Dinan,
  Rockt\"aschel, Kiela, Szlam, and Weston]{urbanek2019light}
J.~Urbanek, A.~Fan, S.~Karamcheti, S.~Jain, S.~Humeau, E.~Dinan,
  T.~Rockt\"aschel, D.~Kiela, A.~Szlam, and J.~Weston.
\newblock Learning to speak and act in a fantasy text adventure game.
\newblock \emph{arXiv preprint arXiv:1903.03094}, 2019.

\bibitem[Vinyals and Le(2015)]{vinyals2015neural}
O.~Vinyals and Q.~V. Le.
\newblock A neural conversational model.
\newblock \emph{arXiv preprint arXiv:/1506.05869}, 2015.

\bibitem[Wang et~al.(2017)Wang, Wang, Li, Xu, Wang, and
  Wang]{wang-etal-2017-group}
J.~Wang, X.~Wang, F.~Li, Z.~Xu, Z.~Wang, and B.~Wang.
\newblock Group linguistic bias aware neural response generation.
\newblock In \emph{Proceedings of the 9th {SIGHAN} Workshop on {C}hinese
  Language Processing}, pages 1--10, 2017.

\bibitem[Yan et~al.(2013)Yan, Guo, Lan, and
  Cheng]{Yan:2013:BTM:2488388.2488514}
X.~Yan, J.~Guo, Y.~Lan, and X.~Cheng.
\newblock A biterm topic model for short texts.
\newblock In \emph{Proceedings of the 22nd International Conference on World
  Wide Web}, pages 1445--1456, 2013.

\bibitem[Yang et~al.(2017)Yang, Zhao, Zhao, Chen, Zhu, Zhou, and
  Cao]{Yang:2017:PRG:3077136.3080706}
M.~Yang, Z.~Zhao, W.~Zhao, X.~Chen, J.~Zhu, L.~Zhou, and Z.~Cao.
\newblock Personalized response generation via domain adaptation.
\newblock In \emph{Proceedings of the 40th International ACM SIGIR Conference
  on Research and Development in Information Retrieval}, pages 1021--1024,
  2017.

\bibitem[Zhang et~al.(2018)Zhang, Dinan, Urbanek, Szlam, Kiela, and
  Weston]{P18-1205}
S.~Zhang, E.~Dinan, J.~Urbanek, A.~Szlam, D.~Kiela, and J.~Weston.
\newblock Personalizing dialogue agents: I have a dog, do you have pets too?
\newblock In \emph{Proceedings of the 56th Annual Meeting of the Association
  for Computational Linguistics (Volume 1: Long Papers)}, pages 2204--2213,
  2018.

\bibitem[Zheng et~al.(2019{\natexlab{a}})Zheng, Chen, Huang, Liu, and Zhu]{P17}
Y.~Zheng, G.~Chen, M.~Huang, S.~Liu, and X.~Zhu.
\newblock Personalized dialogue generation with diversified traits.
\newblock \emph{arXiv preprint arXiv:/1901.09672}, 2019{\natexlab{a}}.

\bibitem[Zheng et~al.(2019{\natexlab{b}})Zheng, Zhang, Mao, and
  Huang]{aaai2020}
Y.~Zheng, R.~Zhang, X.~Mao, and M.~Huang.
\newblock A pre-training based personalized dialogue generation model with
  persona-sparse data.
\newblock \emph{arXiv preprint arXiv:1911.04700}, 2019{\natexlab{b}}.

\end{thebibliography}
\small
\bibliographystyle{abbrvnat}
\end{document}